\documentclass{llncs}
\usepackage[T2A,T1]{fontenc}
\usepackage[utf8]{inputenc}
\usepackage[russian,english]{babel}
\usepackage[final]{graphicx}
\usepackage{epstopdf}
\usepackage[labelsep=period]{caption}
\usepackage[hyphens]{url}
\usepackage{tabularx}
\usepackage{booktabs}
\usepackage{subcaption}
\usepackage{ upgreek }

\usepackage{array,tabulary,adjustbox} 
\usepackage{longtable}  
\usepackage{multirow} 

\newcommand{\keywords}[1]{\par\addvspace\baselineskip
\noindent\keywordname\enspace\ignorespaces#1}

\begin{document}
\title{Size vs. structure in training corpora \\for word embedding models: \\Araneum Russicum Maximum \\and Russian National Corpus}

\titlerunning{Does size matter?}

\author{
Andrey Kutuzov\inst{1}
\and
Maria Kunilovskaya\inst{2}
}

\institute{
University of Oslo\\
\email{andreku@ifi.uio.no},\\
\and
University of Tyumen\\
\email{mkunilovskaya@gmail.com}
}

\index{Kutuzov, Andrey}
\index{Kunilovskaya, Maria}

\maketitle

\begin{abstract}
In this paper, we present a distributional word embedding model trained on one of the largest available Russian corpora: Araneum Russicum Maximum (over 10 billion words crawled from the web). We compare this model to the model trained on the Russian National Corpus (RNC). The two corpora are much different in their size and compilation procedures. We test these differences by evaluating the trained models against the Russian part of the Multilingual SimLex999 semantic similarity dataset. We detect and describe numerous issues in this dataset and publish a new corrected version.
Aside from the already known fact that the RNC is generally a better training corpus than web corpora, we enumerate and explain fine differences in how the models process semantic similarity task, what parts of the evaluation set are difficult for particular models and why. Additionally, the learning curves for both models are described, showing that the RNC is generally more robust as training material for this task.

\vspace{1em}
\end{abstract}

\section{Introduction}\label{sec:intro}
It is a widespread opinion in machine-learning contexts that more data is more beneficial than careful preprocessing and selection of the existing material. Is this true when it comes to representing the meaning of the words as the function of their contexts? This research aims to find out whether and how the type of corpus used to train a word embedding model affects the quality of the resulting word embeddings for Russian. 

Most experimental work on word embeddings is centered around testing available algorithms and their hyperparameters, while the type of corpora used to train the models attracts less attention. At the same time it is reasonable to suggest that the nature of corpus material behind the model should have some bearing on the performance of the latter. 

In this paper, we rigorously compare Araneum Russicum (arguably the largest web-harvested corpus for Russian) against a small but carefully balanced and designed Russian National Corpus (further RNC). We train comparable word embedding models on both corpora and intrinsically evaluate them using the existing semantic similarity and relatedness datasets. Additionally, we reveal some problems in the Russian part of the widely used Multilingual SimLex999 semantic similarity evaluation set, and publish the corrected version.

The paper is structured as follows. In Section \ref{sec:related} we put our research in the context of the previous work. Section \ref{sec:resources} introduces our corpora, the methods used to train  the models, and the gold datasets. Section \ref{sec:results} presents the evaluation results. In Section \ref{sec:error} we analyze and compare typical errors made by the models, and in Section \ref{sec:concl} we conclude.

\section{Related work}\label{sec:related}
Distributional semantic models have been studied and used for decades; see \cite{turney2010frequency} for an extensive review. \cite{Mikolov:2013} introduced the highly efficient \textit{Continuous skip-gram} (SGNS) and \textit{Continuous Bag-of-Words} (CBOW) algorithms for training predictive distributional models, using dense vectors. The so-called \textit{word embedding} models became a \emph{de facto} standard in the NLP world in the recent years, outperforming state-of-the-art in many tasks \cite{baroni2014don}. In the present research, we use the SGNS implementation in the \textit{Gensim} library.

The issue of evaluating distributional semantic models has a long history and is a subject of many discussions (including the special \textit{RepEval} workshop). In the presence of a particular downstream task, it is always better to evaluate the model on this task. However, when training a general-purpose model, one has to rely on intrinsic evaluation, using one of the available gold datasets. The main methods of intrinsic evaluation are numerous, with many available datasets for English. However, in this paper, we limit ourselves to measuring correlation of semantic similarity scores with human judgments, which is also the most established one. For more information on the semantic similarity and relatedness task, especially in Russian context, we refer the reader to \cite{panchenko2016human}.


The issue of the influence of the training corpora on the performance of word embedding models for Russian was raised in \cite{kutuzov_dialog:2015}. Among other, they compared the models trained on the RNC and a randomly sampled corpus of Russian web pages, about an order larger than the RNC. They found out that albeit much smaller, the RNC consistently outperformed the web corpus in the performance of the models trained on it. In this research, we move this even further by using the Araneum Maximum web corpus, which is almost two orders larger than the RNC. Additionally, we carefully analyze the errors of the models and their learning curves. We hypothesize that the types of errors can be different, as it is known that the models trained on corpora of different types can reflect different `semantic landscapes' (see, among others, \cite{kutuzov2015}).

\section{Resources used}\label{sec:resources}
\subsection{Corpora}
The corpora we employ are the \textit{Araneum Russicum Maximum} which is a web corpus of Russian presented in \cite{benko2016very}, and the \textit{RNC}, which is the flagship academic corpus of Russian.  The size of the former corpus is $\approx10 000$ million words, and the size of the latter is $\approx200$ million words. Both corpora were lemmatized and tagged with Mystem \cite{segalovich2003fast}, so that each token was transformed into the `LEMMA\_PoS' representations. Afterwards, the PoS tags were converted to the Universal PoS tagset \cite{petrov_universal}. Functional words, non-alphabetic tokens, punctuation and one-word sentences were removed. These corpora were used to train word embeddings models, as described in the next subsection.

\subsection{The training algorithm}
The models were trained using \textit{Continuous Skipgram} algorithm \cite{Mikolov:2013}, with vector size 600 and a symmetric context window of 2 words to the left and 2 words to the right. In the choice of the window size we considered the known fact that larger windows induce models that are more `associative', while smaller windows induce more `functional' and `synonymic' models, leading to better performance on similarity datasets \cite{goldberg2016primer}. As we are going to evaluate our models on similarity sets, we chose the narrow window of 2.

Low-frequency words were discarded, by using the frequency thresholds of  10 and 400 for the RNC and the Araneum respectively, resulting in the RNC model containing vectors for 173 816 words and the Araneum model containing vectors for 196 465 words. We used 15 negative samples in both cases, and no downsampling (as we removed all the stop words from the corpora beforehand). The sentences in the corpora were shuffled prior to the training to avoid the influence of corpus ordering, and we iterated over each corpus 5 times. 

\subsection{Russian part of Multilingual SimLex999 as the evaluation set}
One of the widely used gold standard sets for testing the ability of distributional models to detect semantic similarity is the SimLex999 \cite{Hill2015}. It was developed to address numerous issues of the previous datasets. We tested our models' performance on the Russian part of \textit{Multilingual SimLex999} evaluation set introduced in \cite{leviant2015separated}. It was created by translating the original English set; the similarity of the resulting word pairs was re-evaluated in a crowd-sourcing effort. We further refer to this set as \textit{RuSimLex999}. It contains word pairs and the corresponding similarity values (for example, \foreignlanguage{russian}{мудрость-ум} `\textit{wisdom-intellect}' 8.23). We tagged the words in the set with the Universal PoS tags, using Mystem and respecting the dataset section the word belonged to (\textit{RuSimLex999} consists of separate sections on nouns, adjectives and verbs).


Despite its popularity, the manual inspection of original \textit{RuSimLex999} revealed numerous technical flaws. Therefore, we produced an improved version of this dataset by resolving the issues noticed. In this paper we report the performance of our models on this corrected set \textit{RuSimLex965} along with the original \textit{RuSimLex999}. The issues addressed are as follows:
\begin{enumerate}
\item 18 duplicate word pairs (one word pair repeated across the dataset): 
\begin{itemize}
\item six pairs with equal scores (e.g. \foreignlanguage{russian}{брать-получать 4.08} `\textit{take-receive}') were deduplicated and only one copy was retained in the revised set,
\item nine duplicate pairs with different scores (e.g. \foreignlanguage{russian}{принимать-отвергать 0/0.69} `\textit{accept-reject}') - we assigned the average score for all duplicate pairs to the unique pair retained in the set,
\item three pairs consisting of the same words in the reversed order with different scores (e.g. \foreignlanguage{russian}{работодатель-работник} 2.08 `\textit{employer-employee}'; \foreignlanguage{russian}{работник-работодатель} 1.77 `\textit{worker-employer}') - we retained one pair with the average score for all duplicate pairs,
\item one pair containing two identical words (\foreignlanguage{russian}{палец-палец 9.92 `\textit{finger-finger}'}; most likely from the English `\textit{toe-finger}') was deleted from the set;
\end{itemize}

\item two pairs which are hardly adequate due to frequency concerns: for example, \foreignlanguage{russian}{рашкуль-уголь} 1.38 (\textit{charcoal-coal}). The first word never occurs in the RNC and is found only five times in the Araneum, which means its IPM is as low as 0.0005. It is unlikely that untrained native speakers are able to quantify the difference in this case consistently and yet it is one of the criteria to be met by a gold standard \cite{Hill2015}. It is not surprising that this word along with \foreignlanguage{russian}{бедствование} (\textit{calamitousness}) is unknown to both our models. All the other words from the dataset are covered by the models;
\item there are several typos in the set. We had to correct spelling in three pairs (\foreignlanguage{russian}{путешествие-за\textbf{во}вание, 0.69; \textbf{приво}ряться-казаться, 4.92; отсутствие-при\textbf{сту}ствие, 0.08})
\item in the pair \foreignlanguage{russian}{мука-горе} 0.277 (\textit{torment-grief}) there is an unnecessary ambiguity: \foreignlanguage{russian}{мука} can mean both `\textit{torment}' and `\textit{flour}' in Russian, depending on the stress. The original English SimLex999 pair is `\textit{agony-grief}', but the correct reading is arguably not the first coming to the mind of the Russian speaker. This pair was deleted from the revised set;
\item many translations could have been more adapted: the Russian test set contains many loan words (\foreignlanguage{russian}{джет, цент, доллар, бренди} `\textit{cent, brandy}') at the same time lacking the respective Russian words \foreignlanguage{russian}{рубль, водка} (`\textit{ruble, vodka}').
\end{enumerate}

It is noteworthy that the original English SimLex999 is free of the above shortcomings. In the German translation, there is one duplicate pair with different scores (\textit{schlecht, schrecklich}); the Italian counterpart had four duplicate pairs, including one with the same scores (\textit{felice, arrabbiato}). 

With PoS consistency in mind, we had to further delete 12 word pairs that were impossible to annotate in accordance with the evaluation set logic (these pairs when tagged include different parts of speech, which might affect the model performance).
In three more pairs we had to adjust lemma tags to  ensure homogeneity of the set. For example, the default out-of-context tagging, which returned the pair \foreignlanguage{russian}{ранить\_VERB-бесстрастный\_ADJ} (literary `\textit{to injure-unemotional}'; most likely from `\textit{fragile-frigid}'), was changed to the intended \foreignlanguage{russian}{ранимый\_ADJ-бесстрастный\_ADJ} (`\textit{vulnerable-unemotional}'), because \foreignlanguage{russian}{ранимый\_ADJ} (`\textit{vulnerable}') was frequent in the tagged corpora. 
After tagging, we also had to fix some lemmatization errors such as \foreignlanguage{russian}{виски} (`\textit{whisky}'), which was lemmatized as \foreignlanguage{russian}{висок} (`\textit{temple}') or \foreignlanguage{russian}{легкое} (`\textit{lungs}') paired with `\textit{liver}', which was lemmatized to the Russian lemma \foreignlanguage{russian}{легкий} (`\textit{easy or light}'). 

Thus, the cleaned and improved \textit{RuSimLex965} semantic similarity evaluation test set includes 965 unique word pairs, consistently PoS-tagged and congruent with the Mystem output. 

\subsection{RuSSE evaluation sets}
For comparison, we also report results for 3 evaluation sets described in \cite{panchenko2016human}:
\begin{enumerate}
\item RuSSE HJ; translated to Russian from the widely used datasets for English;
\item RuSSE WS353 Similarity; translated from the \textit{WS353} dataset \cite{finkelstein2001placing};
\item RuSSE WS353 Relatedness; the same source.
\end{enumerate}
These alternative sets were produced in the context of RuSSE, the first semantic similarity shared task for Russian. In all three datasets, the scores were obtained by a crowd-sourcing initiative employing native Russian speakers.

Unlike \textit{RuSimLex999}, these sets include only nouns with the exception of one pair \foreignlanguage{russian}{подписать-перерыв} (`\textit{to sign}'-`\textit{a break}') 0.0. Other internal discrepancies of the sets include a  pair consisting of two identical words `\foreignlanguage{russian}{тигр-тигр}' (`\textit{tiger}') 0.875, two duplicate pairs `\foreignlanguage{russian}{приспособление-инструмент}' (`\textit{appliance-tool}') 0.708/0.615 and a pair with a non-Cyrillic non-word `\foreignlanguage{russian}{фонд-cd'} (`\textit{fund-cd}') 0. Agian, we PoS-tagged these sets with Mystem.
After filtering and preprocessing the full HJ set contains 394 word pairs, the WS353 similarity component contains 248 pairs and WS353 relatedness component contains 199 pairs. 

Note that the HJ dataset is produced from several different English sets (most pairs come from \textit{WS353}) and does not distinguish between semantic similarity and relatedness. The separate \textit{WS353-sim} and \textit{WS353-rel} test sets are more consistent; however, they are much smaller than \textit{RuSimLex999}, contain only nouns and still suffer from the shortcomings identified in \cite{Hill2015}. It is also important that \textit{RuSimLex999} features significantly higher inter-rater agreement than the RuSSE HJ data set (Krippendorf's alpha 0.57 and 0.49 respectively). Because of these factors, we chose \textit{RuSimLex999} as our primary evaluation measure, despite its flaws described above.


\section{Results}\label{sec:results}
For both models we measured Spearman correlation between the similarities produced by the models and the scores provided in the datasets. In the two cases of out-of-vocabulary word pairs for the original dataset, 0.0 was used as a placeholder for the model similarity. Table \ref{tab:sets} presents the results. All the scores are statistically significant, with \textit{p} value well below 0.01 level.
\begin{table}
\begin{center}
\begin{tabular}{ccc|ccc} 
\toprule
\textbf{Corpus}& \multicolumn{2}{c}{\textbf{SimLex}}&\multicolumn{3}{c}{\textbf{RuSSE sets}}\\ 
& \textit{RuSimLex999} & \textit{RuSimLex965} & \textit{HJ} & \textit{WS353-Sim} &\textit{WS353-rel} \\
\midrule
RNC  & \textbf{43.22} & \textbf{42.55} & 70.69 & 74.11 &  59.07\\
Araneum & 42.52 & 41.46 &\textbf{73.38 }&\textbf{77.51} & \textbf{63.15}\\
\bottomrule
\end{tabular}
\caption{Spearman’s $\rho$ for models scores correlation against gold datasets.}
\label{tab:sets}
\end{center}
\end{table}

The RNC-based models consistently outperform the Araneum-based ones on the SimLex datasets, but lose when evaluated against the RuSSE datasets. We suppose that the reason for this is that the RuSSE sets are not as rigorous in distinguishing semantic relatedness (can the word \textit{a} be substituted with the word \textit{b}?) versus semantic similarity (is the word \textit{a} associated with the word \textit{b}?). \textit{WS353-Sim} contains only similar pairs and \textit{WS353-Rel} contains only related pairs (HJ incorporates both, together with several other smaller test sets).

Consider the pair \foreignlanguage{russian}{`день-рассвет'} (`\textit{day-sunrise}'). In \textit{RuSimLex999}, its score is only 1.54 (rank 655 out of 1 000), thus this pair elements are quite far away from each other. At the same time, in \textit{WS353-rel}, the same pair features score 4.44 and is ranked 67 out of 249, meaning that these two words are really close. In \textit{WS353-sim}, there are no related pairs at all.

Thus, the large Araneum corpus provides better training data to properly rank \textit{either} related or similar pairs. However, when faced with the task to rank similar pairs higher than simply associated ones, it shows up as inferior to the RNC. It means that more data helps only when the downstream task allows to not care for one of the closeness types. If, on  the other hand, one needs to clearly rank the `\textit{coffee-americano}' pair higher than the `\textit{coffee-cup}', smaller but balanced corpora pay off.
Also, the corrected \textit{RuSimLex965} seems to be a bit more difficult for the models than the original one.



\begin{table}
\centering
\begin{tabular}{c|c|c|c}
\toprule
\textbf{PoS} & \textbf{Araneum} & \textbf{RNC} & \textbf{Number of pairs} \\
\midrule
Nouns &41.67 &43.49 & 653  \\
Adjectives &\textbf{47.92} &42.31 & 97  \\
Verbs &44.20 &\textbf{44.65} & 215  \\
\bottomrule
\end{tabular}
\caption{Correlations for different PoS subsets of \textit{RuSimLex965}.}
\label{tab:pos}
\end{table}

If we measure the correlation on separate subsets consisting of pairs belonging to one and the same PoS, some interesting differences can be observed. Table \ref{tab:pos} presents these scores. For nouns and verbs, the RNC model is better, and for both models verbs are easier. Unexpected results show up for adjectives, which seem to be most difficult for the RNC models, but the easiest for the Araneum one: actually, this is the only PoS-subset on which the Araneum model \textit{outperforms} the RNC-based one, for reasons unclear. Note also that the fact that the RNC model is better with nouns on \textit{RuSimLex965} does not prevent it from losing on the RuSSE sets which contain only nouns. This again proves that these sets feature different kind of similarity scores, arguably mixed with relatedness.



\subsection{Learning curves}
\begin{figure}[h]
\begin{center}
\begin{subfigure}{.45\textwidth}
\includegraphics[keepaspectratio,scale=0.3]{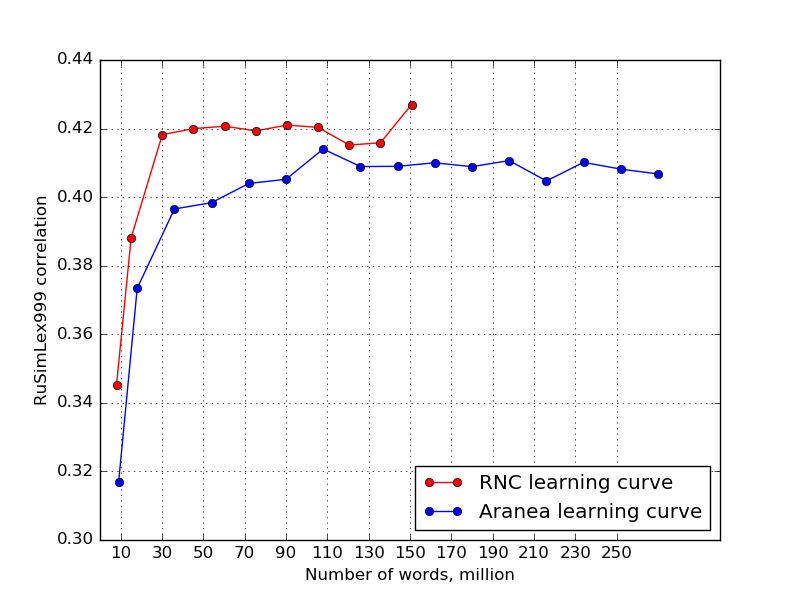}
\caption{RuSimLex999}
\end{subfigure}
\begin{subfigure}{.45\textwidth}
\includegraphics[keepaspectratio,scale=0.3]{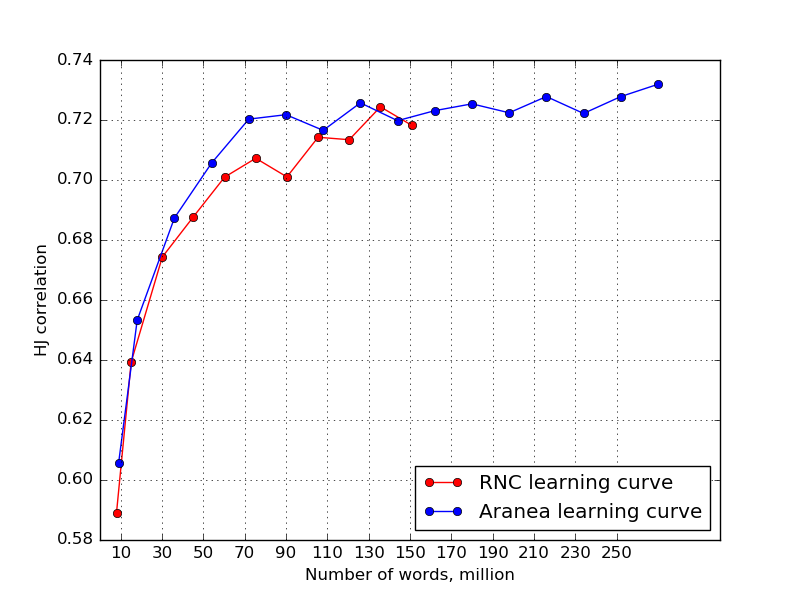}
\caption{HJ}
\end{subfigure}
\begin{subfigure}{.45\textwidth}
\includegraphics[keepaspectratio,scale=0.3]{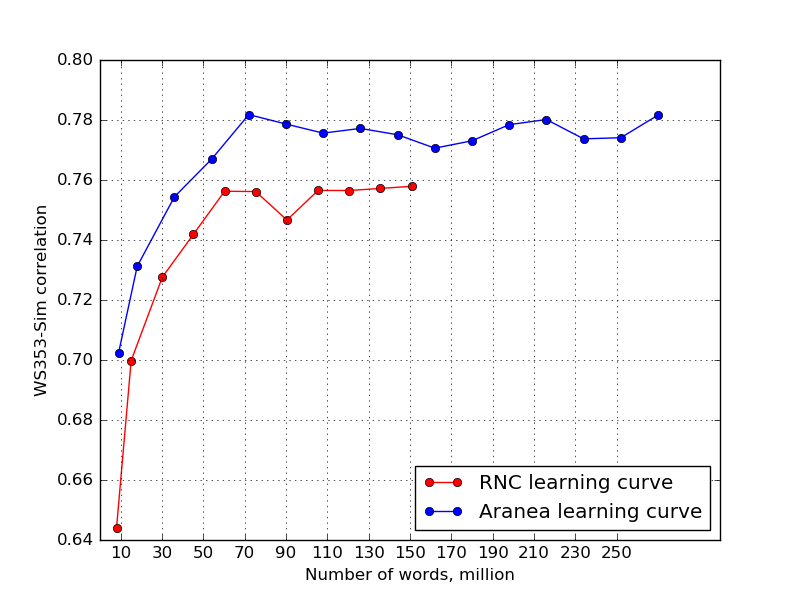}
\caption{WS353 Similarity}
\end{subfigure}
\begin{subfigure}{.45\textwidth}
\includegraphics[keepaspectratio,scale=0.3]{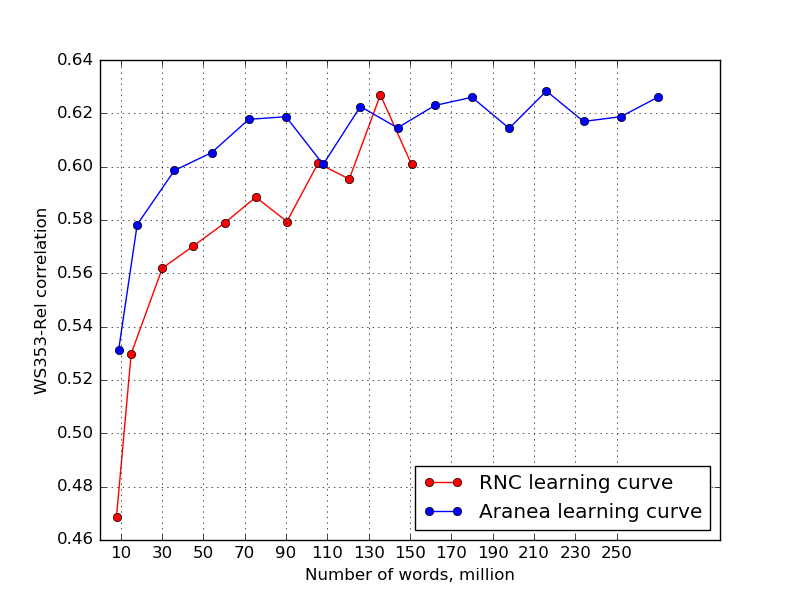}
\caption{WS353 Relatedness}
\end{subfigure}
\caption{Learning curves for Araneum-based and RNC-based models.}
\label{fig:curve}
\end{center}
\end{figure}

We additionally studied how fast the models were in achieving a good performance, as we added more data. To this end, we trained models on the incrementally increased `slices' of our corpora (e.g., the first 10 million words, then the first 20 million words, etc.). For all these models the frequency threshold hyperparameter was set to 10. The results are shown in the Figure \ref{fig:curve}.

For \textit{RuSimLex999}, the RNC provides much better training data from the very start. Even the models trained on the first 10 million words of both corpora already differ by 5 points. Further on, as we train the models on more and more text data, this difference is preserved: the RNC model consistently outperforms the Araneum model, and never vice versa. If we evaluate Araneum trained on approximately the same amount of data as the whole RNC, its \textit{RuSimLex999} performance will achieve only 0.41 (0.43 for the RNC).

The models show similar patterns of development as they are fed with more data. Particularly, moving from 10 to 15..18 million training words makes a huge difference, as well as moving further on to 30 million. After that, the performance on \textit{RuSimLex999} stabilizes and improvements become much smaller.  Another interesting discovery is that sometimes after adding more data, the performance drops for a while. It happens almost simultaneously with the RNC and the Araneum models near 110 million words mark. However, after another 20 million words the models overcome this drop and return to gradual improving.

On the RuSSE test sets, the Araneum models outperform the RNC ones from the very beginning as well. However, except for the \textit{WS353-Sim} set, the results are unstable, with the RNC achieving comparable performance at some times, or even outperforming the Araneum model. Interestingly, on the \textit{HJ} and \textit{WS353-Rel} sets, the RNC model achieves comparable performance at approximately the same moment (about 110 million words) when the Araneum model comes closest to the RNC one on the \textit{RuSimLex999}. This further supports our hypothesis that these sets are complementary. At the same time, \textit{WS353-Sim} should in theory be similar to \textit{RuSimLex999}, but in practice it is the most consistent in showing Araneum outperforming RNC. We have not come to any final conclusion about the reasons for this behavior and leave this for future work.

\section{Error analysis: what the models do wrong}\label{sec:error}





\subsection{Model-specific errors}
To compare performance of the models based on the two corpora, we analyzed each model's distinctive errors against the gold standard and relatively more adequate judgment of the competing model. To arrive at the list of these errors for each model we took the following steps:
\begin{enumerate}
\item ranked pairs by their similarity scores produced by the two models and by human raters; we had to assign pairs with the same score the same ranks;  
\item determined the delta between the ranks of word pairs in the descending list of similarity scores for each model and the gold standard (\textit{RuSimLex965}) (columns 9 and 11 in Table \ref{tab:err});
\item sorted the deltas in the descending order and determined quartiles in the sort: Q1 represents pairs whose model score is very different from the gold one; the lower the quartile, the more accurate the model's assessment is;
\item distinctive errors of each model against each other are pairs with contrasting ranks in the lists above; at their strongest, they come from opposite quartiles. We found the differences between the quartile numbers for each pair and sorted the list in descending order (column 13). 
\end{enumerate}

Table \ref{tab:err} shows the top three and bottom three rows in the results. To make our reasoning explicit let us consider the first example from Table \ref{tab:err}. The rank difference between Araneum and \textit{RuSimLex965} for the hyponym/hyperonym pair \foreignlanguage{russian}{`кот-питомец'} (`\textit{cat-pet}') is twice smaller (234) than for RNC (435). In this case Araneum overestimates similarity placing the pair higher in the ranking (122 is higher than 356), while the RNC underestimates similarity – the pair is way below where it should be from the gold standard point of view (791 instead of around 356). The pair belongs to Quartile 4 (Araneum) and Quartile 1 (RNC) of the respective descending lists of absolute rank differences. The value in column 13 (3) indicates that the RNC similarity estimation for this pair is very distant from the gold standard and Araneum.

\begin{table}
\begin{center}
\resizebox{\textwidth}{!}{%
\begin{tabular}{l|p{75px}|c|c|c|c|c|c|c|c|c|c|c} 
\toprule
\multirow{2}{*}{\textbf{No}} & \multirow{2}{*}{\textbf{Word pair}} & \multicolumn{3}{|c|}{\textbf{Similarity scores}} & \multicolumn{3}{|c|}{\textbf{Ranks}} & \multicolumn{4}{|c|}{\textbf{Rank differences and Quartiles}}& \multirow{2}{*}{\textbf{Diff}}\\ 
& & Araneum & RNC & SimLex & Araneum & RNC & SimLex & $\Updelta$ Ar.-SimLex Ranks & Quart & $\Updelta$ RNC-SimLex Ranks & Quart & \\ 
\midrule
1 & 2 & 3 & 4 & 5 & 6 & 7 & 8 & 9 & 10 & 11 & 12 & 13 \\
\midrule
1. & \foreignlanguage{russian}{кот-питомец} \newline `\textit{cat-pet}'& 0.56 & 0.22 & 0.45
 & 122 & 791 & 356 & 234 & 4 & 435 & 1 & 3 \\
2. & \foreignlanguage{russian}{витамин-железо} `\textit{vitamin-iron}' & 0.38 & 0.23 & 0.27 & 420
 & 773 & 511 & 91 & 3 & 262 & 1 & 2 \\ 
3. & \foreignlanguage{russian}{аргументировать-подтверждать} `\textit{argue - justify}' & 0.35 & 0.24 & 0.55 & 454 & 734 & 270 & 184 & 3 & 464 & 1 & 2 \\   \midrule
\multicolumn{13}{c}{. . .} \\ 
\midrule
963. & \foreignlanguage{russian}{позволять-разрешать} \newline `\textit{allow-permit}' & 0.21 & 0.41 & 0.94 & 767 & 289 & 16 & 751 & 1 & 273 & 3 & -2 \\ 
964. & \foreignlanguage{russian}{скрипка-инструмент} \newline `\textit{violin-instrument}' & 0.32 & 0.44 & 0.45 & 535 & 225 & 356 & 179 & 2 & 131 & 4 & -2 \\ 
965. & \foreignlanguage{russian}{трубка-сигара} \newline `\textit{pipe-cigar}' & 0.33 & 0.44 & 0.53 & 509 & 224 & 284 & 225 & 2 & 60 & 4 & -2 \\
\bottomrule
\end{tabular}}
\caption{Models errors against \textit{RuSimLex965} and the competing model.}
\label{tab:err}
\end{center}
\end{table}

The full content of Table \ref{tab:err} gives a general overview of how the machine judgments compare to each other with respect to distance from the gold standard. The top 12 and bottom 5 word pairs have the interquartile difference of 2 or more. They represent the largest relative errors (given the judgment of the competing model) for the RNC and Araneum respectively. Further top 162 and bottom 165 pairs have the difference of 1.
Note that most of the 965 pairs we tested received the similar scores for Araneum and the RNC, signaling common learning potential of the two corpora. In fact, in our experiment 638 word pairs (66\% of the set) ended up in the same quartile for both models. 

Below we focus on the 101 word pairs (10\%) for which the performance of the models differs the most.
The top 44 pairs (where RNC errs more than Araneum) are dominated by: synonyms (18 pairs), pairs with high levels of association based on contiguity of referents in reality or domain relatedness of the words (14 pairs), and hyponymy/cohyponymy/hyperonymy pairs, particularly verbal (8 pairs).

Interestingly, only 11 pairs out of 44 have a negative rank difference with the gold standard, that is only 11 pairs are placed higher in the ranking than they are in \textit{RuSimLex965}. In two-thirds of the cases the RNC model underestimates similarity in contrast with human scores and almost correct Araneum judgment.

The same semantic groups are prevalent at the bottom of Table \ref{tab:err}, among 57 word pairs for which Araneum gives erroneously higher or lower scores, while RNC gets them almost right (based on rank difference again, not raw scores). The ratio of the groups is more tilted towards unrecognized synonyms, however. Another contrast is in processing strongly associated pairs and different types of hyponymy. While RNC tends to erroneously downplay similarity of concepts related by association or as members of the same classification (general-specific), Araneum places them higher in rank than both the RNC and \textit{RuSimLex965}. The comparison is presented in Table \ref{tab:semerr}.

\begin{table}
\centering
\begin{tabularx}{\textwidth}{cc|c|c|c|p{0.5\textwidth}}
\toprule
\multirow{2}{*}{\textbf{Relation}} & \multicolumn{2}{c|}{\textbf{RNC}} & \multicolumn{2}{c|}{\textbf{Araneum}} & \textbf{Examples \newline (from Araneum model rankings)} \\ 
& under & over & under & over & \\ 
\midrule
synonymy & 16 & 2 & 23 & 2 & \foreignlanguage{russian}{твердый-прочный} `\textit{hard-tough}' \\ 
association & 9 & 5 & 5 & 8 & \foreignlanguage{russian}{кровь-плоть} `\textit{blood-flesh}' \\ 
hyponymy  & 7 & 1 & 2 & 7 & \foreignlanguage{russian}{гитара-барабан} `\textit{guitar-drum}' \\ 
meronymy & 1 & 2 & 2 & 1 & \foreignlanguage{russian}{кость-локоть} `\textit{bone-elbow}' \\
antonymy  & 0 & 1 & 1 & 4 & \foreignlanguage{russian}{работник-работодатель} `\textit{worker-employee}' \\ 
missing value & 0 & 0 & 1 & 0 & \foreignlanguage{russian}{ребячливый-безрассудный} `\textit{childish-foolish}' \\
non-similar & 0 & 0 & 0 & 1 & \foreignlanguage{russian}{дорожка-шар} `\textit{path-ball}' \\ 
\midrule 
\textbf{Total} & 33 & 11 & 34 & 23 &  \\
\bottomrule
\end{tabularx}
\caption{Semantic relations of word pairs with the greatest discrepancy in ranks between each model and \textit{RuSimLex965}/competing model}
\label{tab:semerr}
\end{table}

The analysis shows that both models have difficulties recognizing synonyms. Synonyms constitute by far the largest group of underestimation errors. One of the possible reasons can be simply that there are many synonymic pairs in the test sets, and thus a large portion of them is low-frequency words which means their embeddings are not perfect.

According to \cite{Hill2015}, `similarity is a cognitively complex operation that can require rich, structured conceptual knowledge to compute accurately'. We did not find factors that affected the models’ performance in processing synonyms (such as type of synonymy or their part of speech). Interestingly, two of four pairs whose similarity was overestimated had unreasonably low similarity scores in \textit{RuSimLex965}: \foreignlanguage{russian}{верование-мнение} (`\textit{belief-opinion}') 0.131 and \foreignlanguage{russian}{друг-парень} (`\textit{friend-buddy}') 0.1. This again poses a question of the evaluation set quality.

There are slight differences in how the two models represent association. The model based on Araneum is likely to overestimate similarity in this case, while the RNC model errs on the underestimate side in the same cases. This partly explains high performance of the Araneum models on the \textit{WS353-Rel} test set.


Another group of semantic relations that constitute subtypes of similarity (not association) includes hyponymy/hyperonymy and cohyponymy. 
This is the only single semantic relation with a stark contrast between the two models. For hyponym/hyperonym pairs, the Araneum model learns vectors that return high similarity scores, while this type of similarity goes widely unrecognized by the RNC model (whether this is good, depends on one's particular task, but it seems to be favored by the RuSSE test sets). 

A further finding is that the model trained on Araneum is misled by antonymy. Antonyms are known to share syntactic distributions and therefore, get high semantic similarity scores in distributional models, while humans have no difficulty assigning low similarities to different types of opposites (binaries, gradual/directional, non-binary, relational opposites). Our experiments show that both models overestimate similarity of antonyms, but Araneum performs comparatively worse. All five pairs of antonyms that made it to the list of the Araneum model errors are the so called relative antonyms. Their meanings are opposed given a particular relation between entities or a special situation they are usually part of: \foreignlanguage{russian}{вода-лед, тетя-племянник, работодатель-работник, терять-получать} (`\textit{water-ice}', `\textit{aunt-nephew}', `\textit{employer-employee}',`\textit{lose-gain}'). 

\section{Conclusion}\label{sec:concl}
We compared the performance of Continuous Skipgram word embedding models trained on the very large web corpus of Russian (Araneum Maximum) and the much smaller Russian National Corpus. As our primary evaluation set, we chose the Russian part of Multilingual SimLex999. We revealed numerous flaws in its design and eventually came up with its refined version, \textit{RuSimLex965}. We publish its raw and PoS-tagged variants\footnote{\url{http://rusvectores.org/static/testsets/}}, together with the trained models\footnote{\url{http://rusvectores.org/models/}} and the tagged Araneum Maximum corpus\footnote{\url{http://rusvectores.org/static/rus_araneum_maxicum.txt.gz}}. Note that there are still some conceptual issues in the SimLex999 test set (cf. \cite{oded_goldberg:2016}). Also, all these datasets were originally compiled in English and then re-scored by native Russian speakers, which may degrade their reliability. To address these problems one has to compile a new dataset from scratch, which is outside the scope of the present research. 

With both variants of the evaluation set, our experiments supported the previous work in that a balanced national corpora, albeit smaller, consistently outperform large web-based corpora in semantic similarity evaluation setting. At the same time, the Araneum-based model was superior on the sets containing semantic relatedness scores; thus, this corpus is more suitable for calculating associative, topical and hyponymic relations between words.

Further, we analyzed the speed of performance improving with increasing the size of the training data for both corpora. We show that almost all improvement stops after the first 100 million words for  semantic similarity test sets, and that the RNC `saturates' somewhat faster than the Araneum. For the semantic relatedness test sets, it seems that the model performance does not saturate and continues to improve after this point as well. Finally, we performed an extensive error analysis for both models, revealing typical classes of errors, and how the RNC and Araneum models differ in this respect.

As a future work, we plan to study in more detail how different are RuSSE test sets from the \textit{RuSimLex999}. We also would like to find out whether our findings hold for English and other languages, as well as for other types of intrinsic evaluation (analogical inference, etc).

\bibliographystyle{splncs}
\bibliography{size}

\end{document}